\documentclass[runningheads]{llncs}

\usepackage{textcomp}
\usepackage[T1]{fontenc}
\usepackage{graphicx}
\usepackage{xcolor}
\usepackage{hyperref}
\newcommand{\orcid}[1]{\href{https://orcid.org/#1}{\includegraphics[scale=0.05]{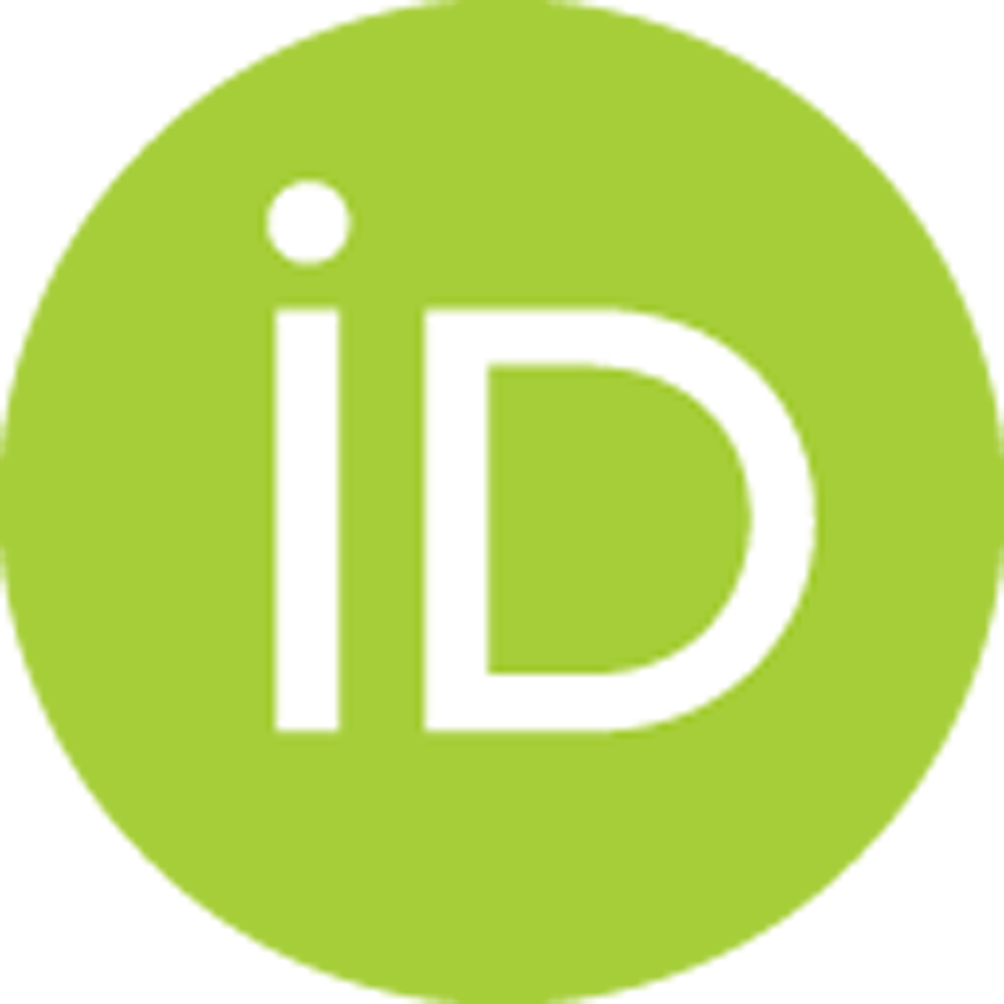}[#1]}}
\usepackage{amsmath}

\begin{document}
\mainmatter

\title{PROFUSEme: \underline{PRO}state Cancer Biochemical Recurrence Prediction via \underline{FUSE}d \underline{M}ulti-modal \underline{E}mbeddings}

\titlerunning{PROFUSEme}

\author{
\underline{Suhang You}\inst{1,2,\orcid{0000-0002-9327-9391}}
\and
\underline{Carla Pitarch-Abaigar}\inst{1,2,\orcid{0000-0002-6015-244X}}
\and
\underline{Sanket Kachole}\inst{1,2,\orcid{0000-0002-1496-2070}}
\and
\underline{Sumedh Sonawane}\inst{3}
\and
Juhyung Ha\inst{3}
\and
Anish Sudarshan Gada\inst{3}
\and
David Crandall\inst{3}
\and
\underline{Rakesh Shiradkar}\inst{1,3,\dag}
\and
\underline{Spyridon Bakas}\inst{1,2,3,4,5,\dag,*,\orcid{0000-0001-8734-6482}}
}

\authorrunning{S.You, et al.}

\institute{Division of Computational Pathology, Department of Pathology and Laboratory Medicine, Indiana University School of Medicine, Indianapolis, IN, USA
\and
Indiana University Melvin and Bren Simon Comprehensive Cancer Center, Indianapolis, IN, USA
\and
Luddy School of Informatics, Computing, and Engineering, Indiana University, IN, USA
\and
Department of Radiology and Imaging Sciences, Indiana University School of Medicine, Indianapolis, IN, USA
\and
Department of Biostatistics and Health Data Science, Indiana University School of Medicine, Indianapolis, IN, USA
\\
\textsuperscript{\dag}: Equal senior authors\\
\textsuperscript{*} Corresponding author: \email{\{spbakas@iu.edu\}}}
\maketitle

\begin{abstract}
    Almost 30\% of prostate cancer (PCa) patients undergoing radical prostatectomy (RP) experience biochemical recurrence (BCR), characterized by increased prostate specific antigen (PSA) and associated with increased mortality. Accurate early prediction of BCR, at the time of RP, would contribute to prompt adaptive clinical decision-making and improved patient outcomes. In this work, we propose \underline{pro}state cancer BCR prediction via \underline{fuse}d \underline{m}ulti-modal \underline{e}mbeddings (\textit{PROFUSEme}), which learns cross-modal interactions of clinical, radiology, and pathology data, following an intermediate fusion configuration in combination with Cox Proportional Hazard regressors. Quantitative evaluation of our proposed approach reveals superior performance, when compared with late fusion configurations, yielding a mean C-index of 0.861 ($\sigma=0.112$) on the internal 5-fold nested cross-validation framework, and a C-index of 0.7107 on the hold out data of the CHIMERA 2025 challenge validation leaderboard.
\end{abstract}


\section{Introduction}
    Almost 30\% of prostate cancer (PCa) patients~\cite{falagario2023biochemical} undergoing radical prostatectomy (RP) experience biochemical recurrence (BCR), characterized by increased prostate specific antigen (PSA). Patients with BCR tend to have higher rates of metastasis and mortality. Early prediction of BCR would allow personalized treatment options and improve patient outcomes. Most patients have magnetic resonance imaging (MRI) prior to RP, post-RP hematoxylin and eosin (H\&E)-stained slides, and documented clinical variables (including age, PSA, Gleason score). All these comprise multi-modal clinically useful data spanning different spatial scales. Prior studies have reported prognostic signatures from uni-modal data, using clinical variables~\cite{cooperberg2011capra}, MRI~\cite{li2021novel}, or digitized H\&E-stained slides~\cite{leo2021computer}.

    In clinical practice, different modalities of data can provide different perspectives about the patient and underlying disease. Combining these for diagnosis and prognosis could lead to better clinical decision-making, which may lead to better outcomes for patients. A recent example of the multi-modal beneficial contribution is the WHO classification of Tumors of the Central Nervous System~\cite{louis20212021} that changed from a pure histologic approach to one integrating molecular profiling with histologic-morphologic assessment for conclusive diagnosis of brain tumors. Leveraging the prime computational capacity, deep multi-modal data fusion is designed to effectively combine extracted multi-modal feature representations, as well as investigate the interplay and complementarity between them. 
    
    Different approaches have been formally described for multi-modal fusion, depending on when feature representations are combined during the processing pipeline~\cite{lipkova2022artificial}. For early fusion, features are directly combined through, for example, concatenation, element-wise sum, or element-wise product of feature vectors. For late fusion, each modality's features are used to train independent models, and the predictions of these models are later ensembled into a single prediction. For intermediate fusion, the complement and interplay between multi-modal feature representations are investigated through their projections into a uniform latent space that eventually provides the best representation.
    
    The Combining HIstology, Medical Imaging (Radiology), and molEcular Data for Medical pRognosis and diAgnosis (CHIMERA) 2025 challenge, conducted in conjunction with the Annual Meeting of the Medical Image Computing and Computer Assisted Interventions (MICCAI) conference, presents a unique paired multi-modal dataset of MRI, digitized H\&E-stained whole slide images (WSIs), and clinical variables for the prediction of BCR in PCa patients. Complementary prognostic information is encoded within these different modalities, with MRI providing a macroscopic overview of the prostate and WSI from RP assessing microscopic tissue heterogeneity.
    
    In this study, we propose the prediction of \underline{pro}state cancer BCR via \underline{fuse}d \underline{m}ulti-modal \underline{e}mbeddings (\textit{PROFUSEme}). Foundation models (FMs) have shown tremendous promise in encoding useful feature embedding representations from medical imaging pixel data. Here, we extract such embeddings from MRI, WSI, and clinical data, which are integrated in a latent space via a cross attention-based intermediate fusion framework, allowing us to learn complementary prognostic BCR information from each of the modalities. More specifically, our approach captures complementary prognostic signals across modalities by first tokenizing patient-wise feature vectors in a shared latent space, then integrating them via a Transformer encoder where self-attention enables cross-modal interactions, and finally mapping the fused representation to a logarithmic risk score and the time to recurrence (TTR) via a Cox Proportional Hazards (CPH) regressor. Quantitative evaluation of our proposed intermediate fusion approach reveals superior performance compared to a late fusion approach using the same feature extraction techniques.

\section{Materials}
    \subsection{Data}\label{data}
        PROFUSEme is developed and evaluated using the training data set of the CHIMERA 2025 challenge\footnote[1] {https://chimera.grand-challenge.org/} and of the LEarning biOchemical Prostate cAncer Recurrence from histopathology sliDes (LEOPARD) 2024 challenge\footnote[2] {https://leopard.grand-challenge.org/}(508 WSI/cases). The CHIMERA 2025 challenge training set consists of 95 patient records, each comprising one tabular clinical dataset, 3 MRI scans of prostate tissue in T2-weighted (T2), apparent diffusion coefficient (ADC), and high b-value (HBV) sequences, and at least one WSI (195 in total). 
        
        For the pathology data censoring and model training splits, we follow the same steps as in our previously described a 2-stage approach, namely ``thinking fast'' and ``thinking slow'' ~\cite{you2024biochemical}. Specifically, in the ``thinking fast'' stage, the auxiliary binary classification of recurrence or not-recurrence is predicted, by setting the time threshold to 22 months and where cases with no BCR and follow-up months smaller than 22 months were excluded. Moreover, we  censored out 8\% of the CHIMERA training data, and 30\% of the LEOPARD data. For the ``thinking slow'' stage, where the TTR is predicted, we used all the CHIMERA training sets WSIs and all LEOPARD challenge training sets.  
        
        From the clinical data, we used 8 attributes and dropped all others. We specifically used: Age at RP, ISUP grade, pathologic T (pT) stage, positive lymph nodes, capsular penetration, positive surgical margins, seminal vesicle invasion, and lymphovascular invasion.

        During training and selection  of the pathology ``thinking fast'' models, we used 5-fold cross-validation  after we randomly selected 20\% of the WSIs from the censored CHIMERA and LEOPARD sets as the held-out test set. From the remaining 80\%, we used 80\% as training data and 20\% as testing data (for an overall training, validation, test split of 64\%:16\%:20\%).
        
        During training and evaluation of the radiology and the pathology ``thinking slow'' models, we used 5-fold nested cross-validation and set the split ratios for training, validation, and testing to 64\%:16\%:20\%, as provided by CHIMERA. All LEOPARD WSIs were included in the training split of the pathology model.
   
    \subsection{Data Pre-processing}
    
        Data pre-processing is depicted in the left of Fig.~\ref{fig:chimera_pipeline} (A). The WSIs were preprocessed, and their tissue masks were segmented based on CLAM~\cite{lu2021data}. Each segmented area was deconvolved to H-E-DAB color space~\cite{ruifrok2001quantification} and pixels of low `E' intensity were discarded. The curated foreground segments were partitioned into non-overlapping patches of about $3.2 mm^2$ (i.e. 224$\times$224, 8mpp, 1.25x magnification for CHIMERA; 448$\times$448, 4mpp, 2.5x magification for LEOPARD) in the ``thinking fast'' stage, and 75\% overlapping patches of about $0.25 mm^2$ (1024$\times$1024, 0.5mpp, 20X magnification, and 512px step size for CHIMERA; 2048$\times$2048, 0.25mpp, 40X magnification, and 1024px step size for LEOPARD) in the ``thinking slow'' stage.
        
        The ADc and HBV MRI sequences were rigidly co-registered using ITK and ITK-Elastix to T2W, which was accompanied by a dedicated tissue mask. During registration, we set the number of spatial samples to 3000, and a fallback sampling to 300 for cases that could fail registration due to oversampling. All MRI scans were then masked by the bounding box of the prostate gland mask (with padding to achieve a square box in the axial plane).

        As described in  Sec.~\ref{data}, clinical data are pre-processed by selecting certain attributes. We processed these attributes differently to adapt to different fusion frameworks.


\section{Method}

    \begin{figure}[th!]
        \centering
        \includegraphics[width=0.85\textwidth]{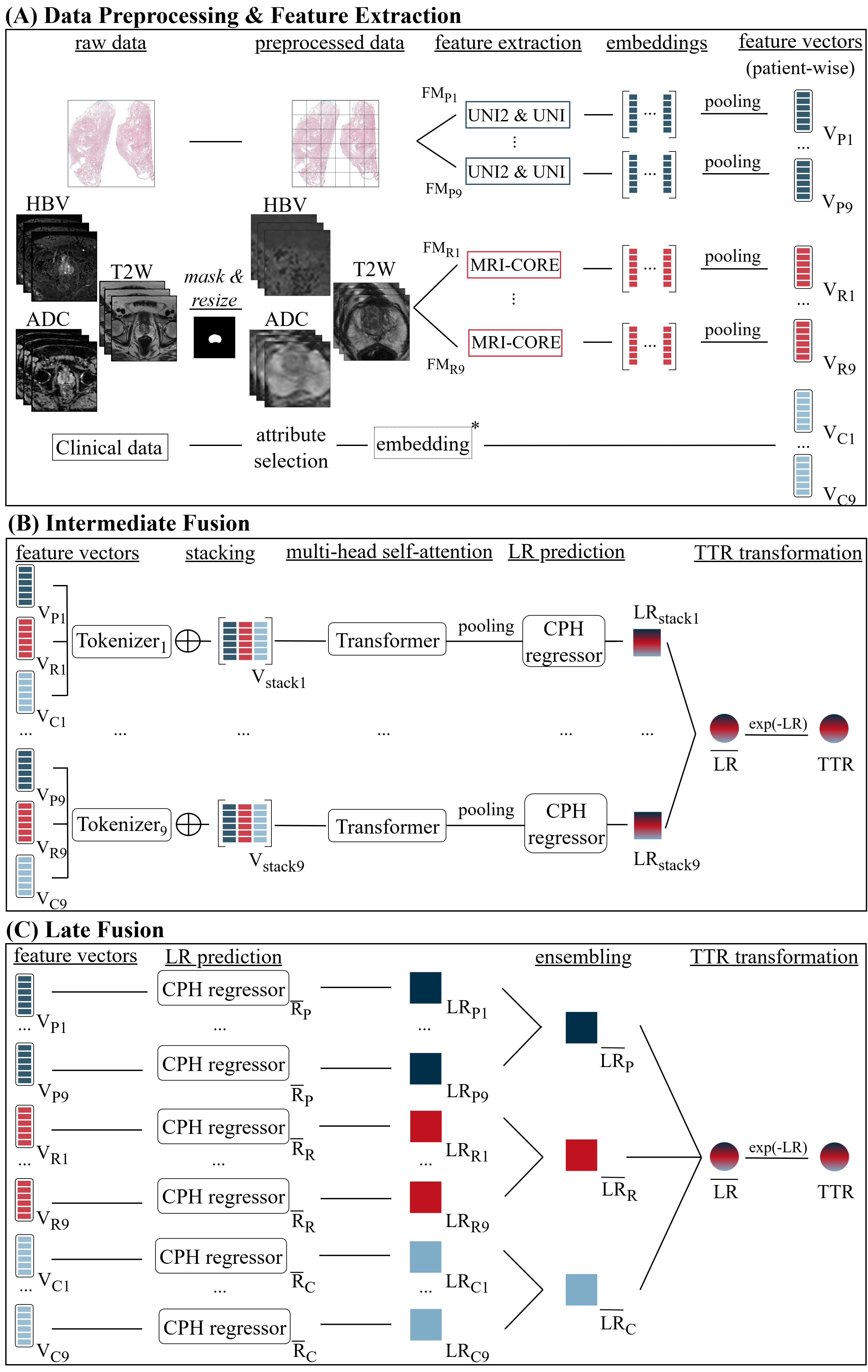}
        \caption{Visual overview of the proposed PROFUSEme framework. (A) Data Preprocessing \& Feature Extraction. The asterisk (*) symbol shows that clinical data are encoded differently in intermediate fusion (in vectors) and late fusion (in dummy variables).
        (B) Intermediate Fusion.
        (C) Late Fusion.
        }
        \label{fig:chimera_pipeline}
    \end{figure}

    PROFUSEme aims at fusing information from three modalities via a tokenizer $\rightarrow$ transformer $\rightarrow$ survival head paradigm. Patient-wise feature vectors from each modality are encoded independently with domain-specific encoders. Modality-specific tokenization layers map each input type into a shared latent space, ensuring that clinical, radiology, and pathology embedding vectors (which differ in their original dimensionality) are represented with a common embedding dimension. These tokenized modality representations are treated as independent tokens that are stacked in the modality dimension to form a matrix input. Learnable positional encodings are added to each modality's token, and the resulting matrix is processed by a compact Transformer encoder. Within this encoder, the multi-head self-attention mechanism performs cross-attention across modalities, enabling the model to capture complementary prognostic information and inter-modality dependencies. Following the Transformer encoder, the output tokens are aggregated using mask-aware mean pooling, ensuring missing modalities do not bias the fused patient-wise representation. This pooled embedding is passed to a linear survival head that outputs a single log-risk score. Our model is trained using the widely adopted `Cox partial log-likelihood loss,' and evaluated using the concordance index (C-index)~\cite{uno2011c} to quantify the model’s ability to rank patients by risk in accordance with observed survival times. For the comparative evaluation with the late fusion approach, we independently built different regressors to predict log-risks based on the CPH regressor and ensembled them to predict TTRs. Our proposed methodological overview is shown in Fig.~\ref{fig:chimera_pipeline}.

    \subsection{Feature Extraction}
    
        Different strategies are used to extract patient-wise pathology feature vectors $V_P$ and patient-wise radiology feature vectors $V_R$ (Fig.~\ref{fig:chimera_pipeline} (A)). The patient-wise clinical feature vector $V_C$ is encoded separately as discussed in Sec.~\ref{feature_pool}

        \subsubsection{Pathology Feature Extraction.}
        
            Patch-wise embeddings from WSIs were extracted following our previous 2-stage framework~\cite{you2024biochemical} with a modification to keep the same physical field of view per patch, irrespective of the different WSI magnifications in the CHIMERA \& LEOPARD datasets. Specifically, we used UNI2~\cite{chen2024uni} for the lower resolution ``thinking fast'' stage feature embedding, and UNI~\cite{chen2024uni} for the higher resolution ``thinking slow'' stage feature embedding. The selected pathology FMs ($FM_{Pi}$, $i = 1,2,3,...,9 $) in the two stages are obtained by experimenting with different combination of FMs between UNI, UNI2, Optimus-0~\cite{hoptimus0}, and Optimus-1~\cite{hoptimus1}. The extracted patch-wise pathology embeddings have a dimension of $M\times1024$, where M is the number of selected patches. To produce the patient-wise pathology feature vector, the extracted patch-wise embeddings were first reduced by a fully connected layer and then pooled (as in our previous work~\cite{you2024biochemical}). We then perform a top k feature selection and self-attention pooling, which results in a pooled pathology embedding per patient (size:$1\times512$). The self-attention was learned during the training of the independent models for late fusion. For intermediate fusion, we directly used these independently trained model weights during the pooling of patch-wise embeddings. 
        
        \subsubsection{Radiology Feature Extraction.}
        
            After data pre-processing, slice-wise embeddings were extracted from the stack of MRI sequences by leveraging the MRI-CORE~\cite{dong2025mri} FM ($FM_{Ri}$, $i = 1,2,3,..., 9$). With an input of $M\times 3 \times W \times H$ stacked sequences (HBV, ADC, T2W concatenated at $2^{nd}$ dimension), the extracted slice-wise embeddings have a dimension of $M\times256\times16\times16$, which was further flattened to $M\times65536$, where M is the number of axial slices, $W$ and $H$ is the width and height of the slice. To create the patient-wise feature vector, the slice-wise embedding was first reduced by two fully connected layers and then pooled using a self-attention mechanism, resulting in a pooled patient-wise feature vector (of size $1\times512$). Similar to pathology feature extraction, self-attention was learned during the training of the independent models for late fusion. In intermediate fusion, we directly used these independently trained weights during the pooling of slice-wise embeddings.  
                    
        \subsubsection{Clinical Feature Encoding.}\label{feature_pool}

            The selected clinical data attributes were encoded differently in intermediate fusion and late fusion. In intermediate fusion, selected attributes were processed according to the data type: categorical attributes were converted to one-hot vectors, where the number of dimensions was dependent on the number of classes, while numerical attributes were Z-score normalized. The selected encoded attributes were then concatenated to form a patient-wise clinical feature vector $V_C$ (of size $1\times25$).
            
            In late fusion, the attributes were similarly processed but not encoded as one-hot vectors since they were not inputs of neural networks. Instead, they were encoded as dummy variables (for simplicity, we still denote the dummy variables as $V_C$). The asterisk in Figure~\ref{fig:chimera_pipeline}(A) notes that the processes are slightly different between intermediate and late fusion during feature extraction; as discussed before, the intermediate fusion requires vectorized input, but the late fusion does not. 

    \subsection{Fusion Approaches}
        
        \subsubsection{Intermediate Fusion Approach.}
        
            Fig.~\ref{fig:chimera_pipeline} (B) depicts the process of intermediate fusion. A patient's multi-modal features consist of a pathology feature vector $V_P$ (size $1\times512$), radiology feature vector $V_R$ (size $1\times512$), and clinical feature vector $V_C$ (size $1\times25$). They were first linearly tokenized into a shared latent space using the same tokenizer, yielding one token per modality. Each token has size $1\times768$. Multi-modal tokens were stacked $V_{stack}$ (size: $3\times768$) and processed by a compact 4-layer Transformer encoder with multi-head self-attention and GELU activations~\cite{vaswani2017attention,hendrycks2016gaussian}. The fused representation was obtained with masked mean pooling and mapped by a single linear head to a log-risk score $LR_{stack}$. The final log-risk $\overline{LR}$ score was the average of 9-fold log-risk scores, and the predicted TTR was calculated by $e^{(-\overline{LR})}$. 

 
        \subsubsection{Late Fusion Approach.}
        
            Fig.~\ref{fig:chimera_pipeline} (C) depicts the process of late fusion. To integrate the three modalities, we computed the median ($\overline{LR}$) of the predicted log-risk ($\overline{LR}_P$, $\overline{LR}_R$, $\overline{LR}_C$) from each of the independent CPH regressors to obtain the final patient-wise prediction. The late fusion is dependent on the different folds' model training, in that we aggregated the models by taking the median (or mean) element-wise across the independent model weights from the 9 folds (the median or mean models step in Tab.~\ref{tab:internal_result}), and used the aggregated model ($\overline{R}_P$, $\overline{R}_R$, $\overline{R}_C$) to predict the log-risk ($LR_{Pi}$, $LR_{Ri}$, $LR_{Ci}$, $i = 1,2,3,...9$) score from different features, where each pooled feature was attention-weighted and dependent on the original model before ensembling. The predicted TTR was calculated by $e^{(-\overline{LR})}$. For the internal data evaluation, we used 5$\times$5 nested cross-validation in Sec.~\ref{paragraph: training}, the aggregated models are calculated from the inner folds' model weights since they share the same hold-out set.

    \subsection{Cox Proportional Hazard (CPH) Regression}
        The CPH regressor with tokenized fusion embeddings in intermediate fusion (Fig.~\ref{fig:chimera_pipeline} (B)) and the final CPH regressor of each independent regression step in late fusion (Fig.~\ref{fig:chimera_pipeline} (C)) are described by a CPH model~\cite{cox1972regression}, and predict the patient risk $R$ for BCR, which is inversely related to TTR. In the CPH model, the risk $R(S) = e^{h(S)}$ is estimated by the linear function of log-risk $\hat{h}_\beta(S) = h_0 + \beta^T \cdot S$, where $S$ is defined as the feature $\{f_1,f_2,...,f_N\}$. The $h_0$ is the baseline logarithmic risk and equals 0 when elements in $S$ are all zero. This $h_0$ holds constant with the same cohort based on the CPH modeling assumption. For the independent clinical modeling, the co-variate coefficients $\beta$ can be directly calculated from fixed embeddings. For the other regressors, during the Cox regression, the weights $\beta$ were optimized
        by minimizing the following negative log partial likelihood through re-parameterization~\cite{katzman2018deepsurv}, defined as:
        \begin{equation}
            l(\beta) = - \sum_{i:e_i = 1} ( \hat{h}_\beta(S_i) - \log \sum_{j:R(t_i)} e^{\hat{h}_\beta(S_j)} ),
            \label{eq:negative_log_partial_likelihood}
        \end{equation}
         where $e_i$ is the event status (BCR: 1, or not: 0) at follow-up $t_i$ (in months), and $S_i$ is the embedding. $R(t_i)$ indicates that the patient, whose inputs are the WSI and MRI, is still at risk of BCR at time $t_i$. In our design, for both the intermediate fusion strategy and the late fusion strategy, we defined a self-attention process for the extracted embedding from the FMs (UNI, UNI2, MRI-CORE) to pool the embeddings. 
        We approximate the TTR using $\exp(-1 \times \log R(S))$, since the logarithmic output risk $\log R(S)$ is inversely related to TTR.

    \subsection{Model Training, Evaluation and Selection}\label{paragraph: training}
        We used the Adam optimizer (learning rate $1\times 10^{-4}$) for the independent radiology and pathology models of late fusion, and the AdamW optimizer (learning rate $1\times 10^{-3}$) for the intermediate fusion model.
        Model training, evaluation, and selection were performed on NVIDIA A100 GPUs. Our source code is based on the CLAM platform, and the inference pipeline is publicly available, under an Apache v.2.0 license, at \url{https://github.com/IUCompPath/CHIMERA}.
    
        For pathology training, we adopted our previous 2-stage (``thinking fast \& slow'') model selection strategy~\cite{you2024biochemical}.
        For the ``thinking fast'' stage, to select the best saved checkpoints of model weights, we used 5-fold cross-validation with a fixed hold-out test set which was randomly chosen from the CHIMERA training set. The model weights with the best validation loss were saved for selection. The best validation and test C-index from 5 saved checkpoints were selected for the framework during training and inference. Model performance was assessed using AUC since this stage is to predict the biochemical recurrence.
        For the ``thinking slow'' stage, during regression, we applied a nested 5-fold cross-validation scheme, without a fixed test set. Each outer fold served as the hold-out test set, while the inner folds were used for model checkpoint selection by choosing the best validation loss. In our setting, 25 censored C-indices are calculated for each parameter setting, and the model parameters with the best mean and standard deviation ($\sigma$) of C-index on outer hold-out were selected. Eventually, we selected the model with features extracted using UNI2 at the ``thinking fast'' step and UNI at the ``thinking slow'' step for fusion experiments. Similarly, during the radiology model and fusion model training, we applied the same nested 5-fold cross-validation without a fixed test set and selected the model checkpoint with the best validation loss for the fusion experiments.
    
        For the model submission to the CHIMERA 2025 challenge, we randomly partitioned the data into 9-fold cross-validation and used all models trained under different folds (as depicted in Fig.~\ref{fig:chimera_pipeline} by the subscript from 1 to 9). The submitted container is applied with the intermediate fusion framework. The final prediction of TTR is calculated by ensembling the predicted logarithmic risk from 9-fold trained model weights. We selected model checkpoints based on early stopping of the validation loss after 250 epochs for the fusion regressors. The epoch threshold was set by calculating the zero-crossing epoch of the second derivative of the training loss curve to avoid under-training.

\section{Results}
    According to our cross-validation results, the best late fusion model is based on the combination of the median of the trained model weights with the average log-risk scores for each model, and at the end using median ensembling across clinical, pathology, and radiology CPH regressor prediction, yielding a mean C-index of 0.797 ($\sigma $= 0.12) (Table~\ref{tab:internal_result}). However, intermediate fusion shows superior performance over all the late fusion combinations, with a mean C-index of 0.861 ($\sigma $ = 0.112) based on cross-validation. This optimal intermediate fusion model in the unseen hold-out CHIMERA validation phase yields a C-index of 0.7107 (validation leaderboard on 2025/08/21)\footnote[3] {https://chimera.grand-challenge.org/evaluation/prostate-cancer-biochemical-recurrence-prediction/leaderboard/}. Based on these results, we submitted the intermediate fusion container for participation in the CHIMERA 2025 Challenge testing phase.

    \begin{table}[t]
        \caption{5-fold nested CV average C-index($\pm\sigma$) for different aggregation combinations of average (AVG) and median (MED) model weights (MW) and predicted log-risk score (LRS) for late fusion. The last two rows are the result of intermediate fusion. (C: Clinical, P: Pathology, R: Radiology)}
        \begin{tabular}{|c|c|c|c|c|}
        \hline
        \textbf{Late Fusion}& \textbf{MED MW} & \textbf{MED MW} & \textbf{AVG MW} & \textbf{AVG MW} \\ 
        \textbf{Models} & \textbf{+ MED LRS} & \textbf{+ AVG LRS}& \textbf{+ MED LRS}  & \textbf{+ AVG LRS}\\
        \hline
        \textbf{C Only}  & 0.814 $\pm$ 0.095 & 0.798 $\pm$ 0.094 & 0.814 $\pm$ 0.095 & 0.798 $\pm$ 0.094 \\
        \textbf{P Only}      & 0.727 $\pm$ 0.223 & 0.734 $\pm$ 0.212 & 0.727 $\pm$ 0.223 & 0.734 $\pm$ 0.212 \\
        \textbf{R Only}       & 0.518 $\pm$ 0.095 & 0.561 $\pm$ 0.078 & 0.524 $\pm$ 0.092 & 0.566 $\pm$ 0.072 \\
        \textbf{C+P}  & 0.780 $\pm$ 0.193 & 0.773 $\pm$ 0.186 & 0.780 $\pm$ 0.193 & 0.773 $\pm$ 0.186 \\
        \textbf{C+P+R (AVG)}   & 0.746 $\pm$ 0.115 & 0.757 $\pm$ 0.108 & 0.752 $\pm$ 0.123 & 0.761 $\pm$ 0.115 \\
        \textbf{C+P+R (MED)} & 0.785 $\pm$ 0.156 & 0.797 $\pm$ 0.120 & 0.788 $\pm$ 0.158 & 0.785 $\pm$ 0.134 \\ \hline\hline
        \textbf{Intermediate} &  \multicolumn{2}{|c|}{\textbf{C+P} } &  \multicolumn{2}{|c|}{\textbf{C+P+R} }\\\cline{2-5}
        \textbf{Fusions} &  \multicolumn{2}{|c|}{0.791 $\pm$ 0.201} &  \multicolumn{2}{|c|}{\textbf{0.861 $\pm$ 0.112}}\\\hline
        \end{tabular}\label{tab:internal_result}
    \end{table}   

\section{Conclusion}
    In this study, we proposed \textbf{\textit{PROFUSEme}} to predict BCR of PCa patients, using an intermediate fusion approach to actively learn interactions across multi-modal information from clinical variables, pre-RP radiology (MRI), and post-RP pathology (WSI) data. Quantitative performance evaluation of PROFUSEme and its comparison with late fusion of independent predictions from each separate modality identified that cross-modality interactions yield superior prognostic value for the prediction of BCR. 
    


    
\section{Code Availability}
The source code of our inference pipeline is available at \url{https://https://github.com/IUCompPath/CHIMERA}.

\section*{Acknowledgements}
    We thank the organizers of the CHIMERA 2025 Challenge (led by Nadieh Khalili, Robert Spaans, and Geert Litjens), for releasing this unique data resource, enabling methodological research in the field of multi-modal data fusion for healthcare.
    
    Research reported in this publication was partially supported by the Informatics Technology for Cancer Research (ITCR) funding program of the National Cancer Institute (NCI) of the National Institutes of Health (NIH) under award number U24CA279629 and the Mike Slive Foundation for Prostate Cancer Research Pilot Grant. The content of this publication is solely the responsibility of the authors and does not represent the official views of the NIH.

\bibliographystyle{splncs04}
\bibliography{bibliography}

\end{document}